\documentclass[compsoc,conference,a4paper,10pt,times]{IEEEtran}
\IEEEoverridecommandlockouts

\usepackage{amsmath,amssymb,amsfonts}
\usepackage{algorithmic}
\usepackage{graphicx}
\usepackage{booktabs}
\usepackage{subcaption}
\usepackage{caption}
\usepackage{mathptmx}

\DeclareMathOperator*{\argmax}{arg\,max}

\begin{document}

\title{Bypassing Backdoor Detection Algorithms in Deep Learning}

\author{\IEEEauthorblockN{Te Juin Lester Tan, Reza Shokri}
	\IEEEauthorblockA{\textit{Department of Computer Science} \\
		\textit{National University of Singapore (NUS)}\\
		\tt{\{lester.tan, reza\}@comp.nus.edu.sg}}
}

\maketitle

\begin{abstract}
	Deep learning models are vulnerable to various adversarial manipulations of their training data, parameters, and input sample.  In particular, an adversary can modify the training data and model parameters to embed backdoors into the model, so the model behaves according to the adversary's objective if the input contains the backdoor features, referred to as the backdoor trigger (e.g., a stamp on an image).  The poisoned model's behavior on clean data, however, remains unchanged.  Many detection algorithms are designed to detect backdoors on input samples or model parameters, through the statistical difference between the latent representations of adversarial and clean input samples in the poisoned model.  In this paper, we design an adversarial backdoor embedding algorithm that can bypass the existing detection algorithms including the state-of-the-art techniques.  We design an adaptive adversarial training algorithm that optimizes the original loss function of the model, and also maximizes the indistinguishability of the hidden representations of poisoned data and clean data. This work calls for designing adversary-aware defense mechanisms for backdoor detection.  
\end{abstract}

\section{Introduction}

Deep learning models are capable of learning complex tasks with a high predictive power.  They are, however, vulnerable to many privacy and security attacks that exploit their large capacity.  The models are susceptible to adversarial manipulations of their training set, parameters, and inputs.  The attacker can degrade the model's predictive power, or change its behavior according to the adversary's objective, by poisoning the training set~\cite{chen2017targeted, shafahi2018poison, gu2017badnets, yang2017generative}, or its parameters~\cite{bagdasaryan2018backdoor}.  It is also possible to adversarially and stealthily manipulate a normal data point to confuse the model into making a wrong prediction~\cite{szegedy2013intriguing}.  

The models can also leak a significant amount of information about their training data, parameters, and inputs.  There is a large body of research on various types of privacy attacks and countermeasures against them, under black-box and white-box access settings, and centralized and distributed learning settings.  Having access to the model predictions or parameters, an adversary can infer sensitive information about the model's training set~\cite{shokri2017membership, nasr2019comprehensive}.  The model predictions can also be exploited to reconstruct the model parameters~\cite{tramer2016stealing}, or its input~\cite{fredrikson2014privacy}.  

\begin{figure}[t]
	\begin{center}
		\centering
		\includegraphics[width=0.4\linewidth]{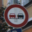}
		\hfill
		\includegraphics[width=0.4\linewidth]{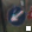}
		\caption{Examples of images with (right) and without (left) a backdoor trigger.  The poisoned model will recognize the trigger and acts adversarially.}
		\label{fig:example_backdoor_image}
	\end{center}
\end{figure}

In this paper, we focus on active attacks against machine learning algorithms.  We specifically focus on a class of attacks known as backdoor attacks, where the adversary manipulates training data and/or the training algorithm and parameters of the model in order to embed an adversarial (classification) rule into the model~\cite{chen2017targeted,gu2017badnets}.  The model behaves normally on all inputs, except for the inputs that contain the adversary's embedded pattern, called the backdoor trigger.  Several types of backdoor triggers have been explored in previous studies. These include input-instance triggers where the backdoor instances correspond to specific inputs in the input space, or pixel-pattern triggers that contain a specific pixel pattern, e.g., the images that contain a stamp, and also semantic triggers where the backdoor instances contain a specific high-level feature, e.g., objects with a particular shape or color.  Figure~\ref{fig:example_backdoor_image} shows an example of an image with a backdoor trigger.  Given the wide range of deep learning applications, backdoor attacks have the ability to cause serious damage, from bypassing facial recognition authentication systems~\cite{chen2017targeted}, to fooling driverless vehicles to misclassify stop signs~\cite{gu2017badnets}. 

A number of backdoor detection algorithms are designed for deep learning~\cite{liu2018fine, chen2018detecting, tran2018spectral, chen2017targeted, liu2017neural}.  These algorithms focus on identifying which inputs contain backdoor, and which parts of the model (its activation functions specifically) are responsible for triggering the adversarial behavior of the model.  For a given adversary model, the detection algorithms try to identify the signatures of the backdoors in the hidden layers of the model, in order to distinguish inputs with the backdoor trigger from clean benign inputs.  Note that the backdoor rule is an exception in the main task represented by the machine learning model.  Thus, to learn the adversarial task along with the main task, the learning algorithm tries to minimize the conflict between the two.  This is what the stat-of-the-art detection algorithms rely on. These algorithms compute various types of statistics on the latent representations of inputs, which can help the defender to separate adversarial and benign data, relying on the distinguishable dissimilarity between the distribution of their latent representations. 

The common implicit assumption of prior defense techniques is that the adversary is unaware of the detection algorithm.  Ignoring adaptive attack algorithms is the main limitation of defense methods in adversarial machine learning.  In the case of adversarial examples, it has been shown that a large number of defense mechanisms can be bypassed by an adaptive attack, for the same weakness in their threat model~\cite{athalye2018obfuscated, carlini2017towards, carlini2017adversarial}.  In this paper, we design an {\bf adversarial backdoor embedding} algorithm for deep learning, that maximizes the latent indistinguishability between adversarial inputs and benign inputs.  We show that the attack strategy can be tailored to any particular detection algorithm and the statistics that the defender uses for identifying backdoors.  We go beyond bypassing specific algorithms.  To be effective against generic detection algorithms, we maximize the latent indistinguishability of input data, using \emph{adversarial regularization}.   

In our threat model, the adversary is capable of exploiting the training algorithm.  We rely on data poisoning and adversarial regularization in our backdoor embedding attack.  We construct a discriminator network which optimizes for identifying any difference between the benign and adversarial data in the hidden layers of the model.  The objective function for the classification model is adversarially regularized to maximize the loss of the discriminator (bypassing network).  Thus, the final model is not only accurate on classifying benign data points according to their clean label, and is accurate adversarial data points according to their adversarial label, but also has indistinguishable latent representation for data points in these two sets.  This enables the compromised model to bypass the detection algorithms which cluster and separate the latent representations of benign and adversarial inputs. 

Our adversarial embedding attack successfully evades several state-of-the-art defenses. As the baseline, for a VGG model trained on the CIFAR-10 dataset, the dataset filtering defense using spectral signatures \cite{tran2018spectral} is able to bring down the backdoor attack success rate of a compromised model to 1.5\%, assuming a static attack strategy (assumed in \cite{tran2018spectral}).  But, a model compromised with our adversarial backdoor embedding algorithm is able to retain an attack success rate of 97.3\%, against the detection algorithm. The dataset filtering defense using activation clustering \cite{chen2018detecting} is similarly able to bring down the static attack success rate of a compromised model to 1.9\%.  But our adversarial embedding algorithm retains a 96.2\% attack success rate, against the detection algorithm.  Feature pruning \cite{wang2019neural} is able to effectively select neurons to prune for a model, and is able to completely remove the backdoor behavior with almost no loss in model accuracy, assuming the baseline static attack.  However, for a model with adversarial embedding, the full removal of the backdoor behavior simultaneously degrades the model accuracy significantly down to 20\% (where 10\% is the random guess).  Thus, all existing detection algorithms fail against the adversarial backdoor embedding algorithm. 

\section{Prior Backdoor Detection Algorithms}

We denote the input space of the model as $X$, where each input instance $x$ comes with its corresponding class label $y$. Backdoor attacks are defined by a backdoor trigger, which is a property $B$ on each input in the input space, such that $\forall x \in X$, $\ B(x)$ is classified as $y_{t} \neq y$, where $y_{t}$ is a target label of the attacker's choice.  We will refer to the set of input instances with $B$ as backdoor instances.

Consider two input-label pairs, (i) a clean input instance $x_{c}$ and its corresponding label $y_{c}$, as well as (ii) a backdoor input instance $x_{b}$ that has a true class label $y_{b}^{true}$, but is wrongly classified as the target label $y_{b}=y_{c}$ due to the presence of the backdoor. Even though both input instances have the same target label, $x_{c}$ contains high-level features corresponding to its true class $y_{c}$, while $x_{b}$ contains high-level features corresponding to both its true class $y_{b}^{true}$ and the backdoor trigger.

A hidden layer in a deep learning network can be treated as the model's latent representation of the input instance, with the neurons in the layer representing different high-level features of the input instance. Given the high-level features present in $x_{c}$ and $x_{b}$ differ, one expects the respective latent representations of these input instances to also differ considerably. Several studies have successfully leveraged this difference in latent representations to detect, or to mitigate the backdoor behavior. While any hidden layer can be treated as the latent representation of the inputs, the defenses are typically applied to the penultimate layer, since it represents the highest-level features extracted by the model.

The proposed defenses we analyze fall into two main categories. The first category of defenses, given a poisoned model, uses the model's latent representations of clean and poisoned instances to determine neurons to prune, in order to remove the backdoor adversarial rule from the network. The second category of defenses uses the latent representations to filter the training dataset, in order to remove most, if not all, of the maliciously injected poisoned samples. The model can then be retrained on the remaining samples to obtain a functional classifier without the backdoor behavior.

\subsection{Feature Pruning}

Wang et al., 2019~\cite{wang2019neural} formulate a detection technique that assuming a known subset of clean inputs, detects possible backdoors and removes them. The authors design a reverse-engineering process based on an optimization function that finds the minimum perturbation required to cause the misclassification of all inputs to a particular target class. This process is applied to every class in the task, yielding a candidate backdoor trigger for each class. Then, based on the intuition that a backdoor trigger is a small perturbation on the input instances, outlier detection based on the median absolute deviation is performed to detect abnormally small perturbations, which are highly likely to be the injected backdoor triggers.

Wang et al. then propose a pruning algorithm that utilizes the reverse-engineered backdoor trigger to remove the backdoor adversarial rule from the model. It does so by recording the mean activation of each neuron $n$ in the hidden layer over clean inputs, $\overline{z_{c}^{n}}$, and over inputs with the backdoor trigger, $\overline{z_{b}^{n}}$. Then, neurons are pruned in the order of decreasing absolute difference in the means:
\begin{equation}
\argmax_{n\in N} |\overline{z_{c}^{n}} - \overline{z_{b}^{n}}|. \label{eqn:pruning_metric}
\end{equation}

The pruning is terminated when the backdoor behavior is fully removed from the model. This defense mechanism assumes that the backdoor adversarial rule in the model is implemented by a large change in activation for the neurons that represent the backdoor features.

\subsection{Dataset Filtering by Spectral Signatures}

Tran, Li, and Madry, 2018~\cite{tran2018spectral} propose a technique based on robust statistics to identify and remove poisoned data samples from a potentially compromised training dataset. First, a network is trained using the poisoned training dataset. For each particular output class label, all the input instances for that label are fed through the network, and their latent representations are recorded. Singular value decomposition is then performed on the covariance matrix of the latent representations, and this is used to compute an outlier score for each input. The inputs with the top scores are flagged as poisoned inputs, and removed from the training dataset. The authors show that this defense succeeds when the means of the latent representations of clean inputs are sufficiently different from the means of the latent representations of the inputs that contain the backdoor trigger.

\subsection{Dataset Filtering by Activation Clustering}

Chen et al., 2018~\cite{chen2018detecting} devise a defense that relies on clustering the latent representations of the inputs. For all input instances the model classifies as a particular class label, the latent representations of the inputs are recorded. Dimensionality reduction is performed using independent component analysis to reduce the recorded latent representations to 10 to 15 features, and k-means clustering is then performed to separate the transformed data into 2 clusters. This clustering step assumes that when projected onto the principal components, the latent representations of the backdoor and clean instances form separate clusters due to the model extracting different features from them.

K-means clustering is instructed to produce 2 clusters, regardless of whether poisoned samples are present.  Chen et al. recommend a process called exclusionary reclassification to determine which of the clusters, if any, is poisoned.  A new model is trained using all training samples excluding one of the clusters. Then, the newly trained model is used to classify the input instances in the cluster, to detect if one cluster is poisoned. 

\section{Adversarial Backdoor Embedding Attack}

Our objective is to construct adversarially poisoned deep learning models that are not detectable using the class of backdoor detection algorithms that try to separate clean and adversarial inputs, which contain backdoor triggers, from their latent representations in the model. 

The defenses above perform well since a significant difference in distribution of latent representations in backdoor instances and clean instances tends to emerge when a poisoned model is trained naively by the attacker. However, the defenses above fail to consider that a sophisticated attacker is able to make the model robust to them by {\bf minimizing their difference in latent representations}. To do so, we introduce a secondary loss function to the training objective function
\begin{equation}
\mathcal{L}(f_{\theta}(x), y) + \mathcal{L}_{rep}(z_{\theta}(x))
\end{equation}
where $x$ is the input instance, $y$ is the target label, $\theta$ is the parameters of the network, $f_{\theta}(x)$ is the class prediction of the network for input $x$, and $z_{\theta}(x)$ is the latent representation of $x$, extracted by the network. $\mathcal{L}_{rep}(z_{\theta}(x))$ represents an additional penalty term that penalizes the model when the distributions of the network activations differ greatly between clean and backdoor inputs. This additional penalty can be tailored to a specific defense that the attacker anticipates, or can be a general penalty that mitigates various defenses, as we will demonstrate. Through the double objective function, the attacker aims to achieve high classification accuracy of the model, while setting certain constraints on the latent representations of the inputs in order to bypass potential defenses.

\begin{figure}[t]
	\centering
	\includegraphics[width=0.8\linewidth]{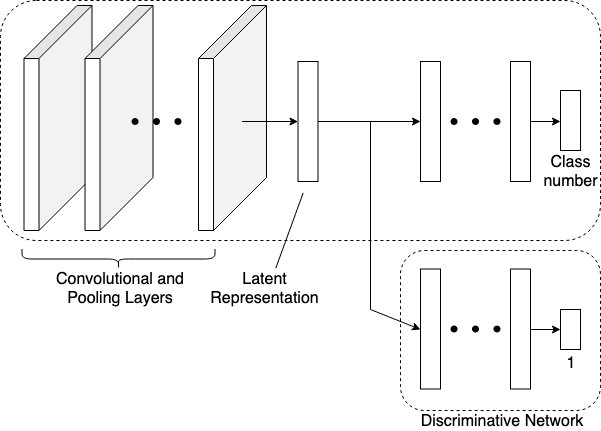}
	\caption{The architecture of our adversarial embedding attack. A discriminator is included that takes the latent representation from the model as input and decides if it is from a backdoor or a clean input.}
	\label{fig:gan_architecture}
\end{figure}

\subsection{Targeted Adversarial Embedding}

We will explore an example of an attacker using the double objective function above to mitigate a specific defense. Consider the pruning defense by Wang et al.~\cite{wang2019neural}, which selects neurons to prune based on the absolute difference in mean neuron activations between clean and backdoor input instances, as presented in Equation \ref{eqn:pruning_metric}.

In order to prevent the backdoor neurons from being selected for pruning, the attacker has to minimize $|\overline{z_{c}^{n}} - \overline{z_{b}^{n}}|$ for each neuron in the backdoor neuron set $N_{b}$. We notice that for any neuron $n$,
\begin{equation}
|k\overline{z_{c}^{n}} - k\overline{z_{b}^{n}}| = k|\overline{z_{c}^{n}} - \overline{z_{b}^{n}}| \leq |\overline{z_{c}^{n}} - \overline{z_{b}^{n}}|
\end{equation}
for any $0 < k < 1$. Thus, by scaling the average activation of neuron $n$ with a sufficiently small $k$, we can make the defense prioritise other neurons for pruning. We thus construct our double objective function to scale the activations in the latent feature representation
\begin{equation}
\mathcal{L}(f_{\theta'}(x), y) + \lambda\mathcal{L}_{rep}(z_{\theta'}(x), z_{target}(x))
\end{equation}
where
\begin{equation*}
z_{target}^{n}(x) = 
\begin{cases}
k \cdot z_{\theta}^{n}(x),& \text{if } n\in N_{b}\\
z_{\theta}^{n}(x),& \text{otherwise}
\end{cases}
\end{equation*}
$\lambda$ is the regularization constant, $\theta$ is the parameters of the naively poisoned model, and $\theta'$ is the parameters of the new model. In order to obtain the set of backdoor neurons $N_{b}$, the pruning defense is first run on the naively poisoned model, and the set of neurons pruned by the algorithm (assuming the defense succeeds) will then contain mostly backdoor neurons. The additional loss function aims to preserve the pathways of the signals in the model, but scale the amplitudes of the signals passing through any neuron in $N_{b}$ by $k$. We use the mean square error as our loss function $\mathcal{L}_{rep}$.

\subsection{Adversarial Embedding}

The objective function proposed above provides a way for the attacker to evade one defense, but it might not transfer well to other defenses. We denote the distribution of the latent representations of clean inputs as $p_{c}$, and the distribution of the latent representations of backdoor inputs as $p_{b}$. Defenses that work on the latent representations of the model, regardless of detection technique, assume the model has learned differences between the distributions $p_{c}$ and $p_{b}$. These differences then inform the defense what inputs are backdoor inputs, or what neurons correspond to backdoor features. Thus, a general form of adversarial embedding will be the one that minimizes this difference such that $p_{c} \approx p_{b}$, to prevent any significant difference from being picked up by the defenses.

We utilise an adversarial network regularization setting (Figure \ref{fig:gan_architecture}), where we treat the initial layers up to the latent representation as a network $H$ that outputs a latent representation given an input. Thus, we have $z_{\theta}(x)=H(x)$. The layers after the latent representation layer form the classification network $C$ that maps from a latent representation to a class probability distribution. The model, then, is the composition $C$ and $H$, i.e. $f_{\theta}(x) = C(H(x))$. We also construct a discriminative network $D$, that maps each latent representation $H(x)$ to a binary classification representing whether the latent representation belongs to a clean or backdoor input.

We then introduce the cost functions for $D$ into our original objective function, obtaining
\begin{equation}
\mathcal{L}(f_{\theta}(x), y) - \lambda\mathcal{L}_{D}\Big(D\big(H(x)\big), B(x)\Big) \label{eqn:adversarial_loss}
\end{equation}
and an objective function for the discriminative network
\begin{equation}
\mathcal{L}_{D}\Big(D\big(H(x)\big), B(x)\Big) \label{eqn:discriminator_loss}
\end{equation}
where
\begin{equation*}
B(x) = 
\begin{cases}
1,& \text{if } x\in X_{b}\\
0,& \text{otherwise}
\end{cases}
\end{equation*}

Thus, the objective of our network is to generate accurate class predictions, and at the same time extract latent representations that the discriminator is unable to classify well as clean or poisoned. We use the cross entropy loss for $\mathcal{L}_{D}$. As our training converges, we expect the distributions of latent representations for clean inputs $p_{c}$ and backdoor inputs $p_{b}$ to converge, such that $p_{c} \approx p_{b}$, thus minimizing any dissimilarities that the defenses rely on to detect backdoor behavior.

\section{Evaluation}

\begin{table}[t]
	\centering
	\begin{tabular}{llcccc}
		\toprule
		Model & Dataset & Tuning LR & Epochs & $k$ & $\lambda$ \\
		\midrule
		DenseNet & GTSRB & $10^{-1}$ & $30$ & $10^{-5}$ & 1 \\
		VGG & CIFAR-10 & $10^{-1}$ & $30$ & $10^{-6}$ & 1 \\
		\bottomrule
	\end{tabular}
	\caption{Hyper-parameters used in the targeted models.}
	\label{tab:attack_parameters_pruning}
\end{table}

\begin{table}[t!]
	\centering
	\begin{tabular}{llcccc}
		\toprule
		Model & Dataset & Epochs & Tuning LR & Discrim. LR & $\lambda$\\
		\midrule
		DenseNet & CIFAR-10 & 30 & $10^{-4}$ & $10^{-3}$ & $50$\\
		DenseNet & GTSRB & 30 & $10^{-4}$ & $10^{-3}$ & $30$\\
		VGG & CIFAR-10 & 1000 & $10^{-3}$ & $10^{-3}$ & $10$\\
		VGG & GTSRB & 30 & $10^{-4}$ & $10^{-3}$ & $20$\\
		\bottomrule
	\end{tabular}
	\caption{Hyper-parameters of our embedding attack.}
	\label{tab:attack_parameters}
\end{table}

\begin{table}[t!]
	\centering
	\begin{tabular}{ccl}
		\toprule
		Layer & Channels & Activation\\
		\midrule
		FC & 256 & Leaky ReLU, Batchnorm\\
		FC & 128 & Leaky ReLU, Batchnorm\\
		FC & 1 & Softmax\\
		\bottomrule
	\end{tabular}
	\caption{Discriminator architecture used in adversarial embedding attack. FC stands for a fully-connected layer.}
	\label{tab:discriminator_architecture}
\end{table}

Our objective is to empirically show the effectiveness of adversarial backdoor embedding attack in bypassing detection algorithms on benchmark deep learning tasks.  

\subsection{Setup}

We perform the above defenses on both a naively poisoned model, as well as a model trained with adversarial embedding, in order to compare the effectiveness of the defenses when facing a sophisticated attacker.  We follow the setting similar to that of the related work, where the detection algorithms are proposed. 

\subsubsection{Datasets}

We perform our evaluation on 2 image classification datasets, to show that our attack is transferable across different applications. We use the CIFAR-10 \cite{krizhevsky2009learning} dataset which consists of 60,000 $32 \times 32$ color images, equally distributed amongst 10 mutually exclusive classes. Of these images, 50,000 are used for training, and 10,000 are used for testing. We also perform our experiments on the GTSRB \cite{Stallkamp-IJCNN-2011} dataset of German traffic signs. It consists of over 51,839 color images, whose dimensions range from $15\times 15$ to $250 \times 250$ pixels (not necessarily square). The images are unevenly distributed amongst 43 mutually exclusive classes. Of these 51,839 images, 39,209 are used for training, and 12,630 are used for testing. Due to the varying sizes of the images, the images are resized to $32 \times 32$ before being passed to the model for classification.

\subsubsection{Models}

We perform our experiments on 2 state-of-the-art deep convolutional neural networks. Since we resize the GTSRB images to the same dimensions as the CIFAR-10 images, we are able to use the same model architecture for both datasets. The model architectures we use are the DenseNet-BC\cite{huang2017densely} model architecture with $L=100,\ k=12$, and the Configuration 'E' VGG \cite{simonyan2014very} architecture with 19 weight layers.

\subsubsection{Backdoors}

We inject a backdoor trigger of a white $4\times4$ square at the bottom right corner in 5\% of training samples, and set their labels to our arbitrarily chosen target label $y_{t} = 2$. We train a model with this poisoned data, obtaining a classifier that contains the backdoor. This poisoned model forms our baseline for comparing the various defence techniques outlined above. We train the models for 100 epochs with an initial learning rate of 0.1, which is divided by 10 at the 50th and 75th epochs.

\begin{figure}[t]
	\centering
	\includegraphics[width=\linewidth]{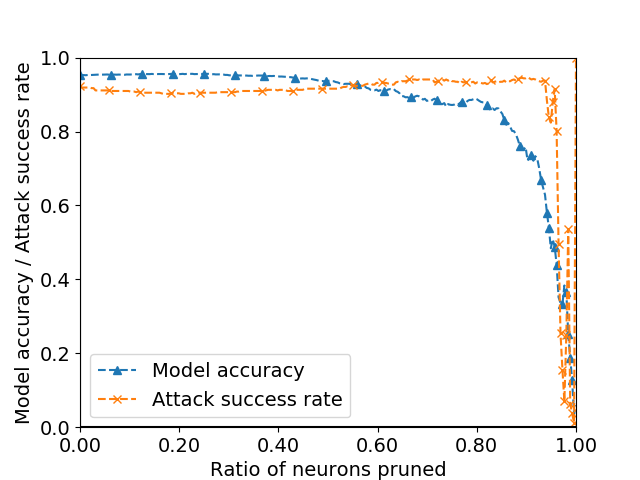}
	\caption{Attack success rate and model accuracy versus the ratio of neurons pruned when pruning \protect{\cite{wang2019neural}} is performed on the baseline model (VGG, GTSRB). Pruning fails to remove the attack success rate without a significant loss in model accuracy.}
	\label{fig:prune_gtsrb_vgg_naive}
\end{figure}

\begin{figure}[t]
	\begin{subfigure}{\linewidth}
		\centering
		\includegraphics[width=\linewidth]{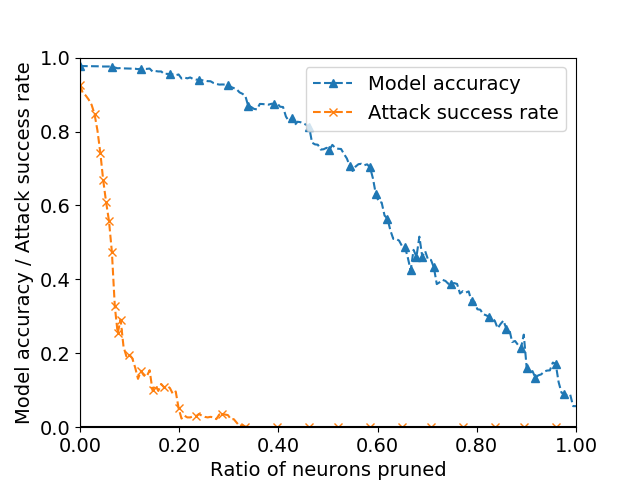}
		\subcaption{Baseline model. Pruning successfully brings the attack success rate down to 0\% with a small trade-off in model accuracy.}
		\label{fig:prune_gtsrb_densenet_naive}
	\end{subfigure}
	
	\begin{subfigure}{\linewidth}
		\centering
		\includegraphics[width=\linewidth]{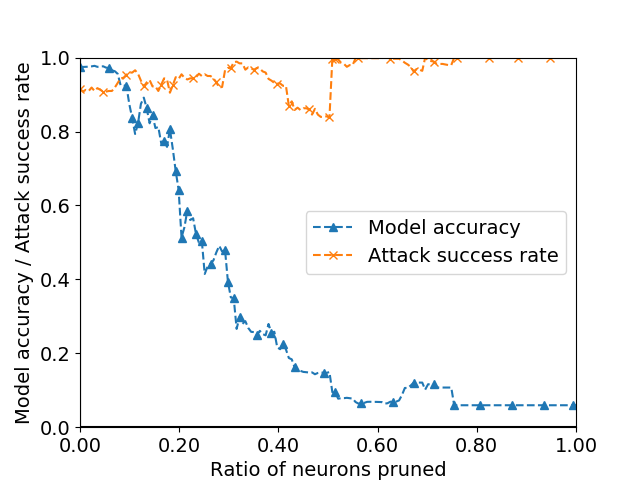}
		\subcaption{Model with targeted adversarial embedding. Pruning is unable to bring down the attack success rate}
		\label{fig:prune_gtsrb_densenet_aware}
	\end{subfigure}
	\caption{Attack success rate and model accuracy versus the ratio of neurons pruned when pruning \protect{\cite{wang2019neural}} is performed (DenseNet, GTSRB).}
\end{figure}

\subsection{Results for Targeted Embedding}

We first perform our targeted adversarial embedding on the baseline network, to evade the feature pruning defense. We assume the reverse-engineering process in the defense perfectly reconstructs the backdoor trigger, and thus we use the actual trigger described above. In order to obtain the set of backdoor neurons, the pruning defense is first performed on the baseline model, until the model accuracy falls by 8\%. The pruned neurons are treated as the backdoor neurons, assuming the defense succeeds in removing the backdoor behavior. The activations of the obtained backdoor neurons are then scaled by $k$. To evaluate the pruning defense, we measure the model accuracy as well as the attack success rate versus the ratio of neurons in the hidden layer that are pruned. The attack success rate is simply the percentage of backdoor inputs in the test set the model classifies as the backdoor target label. For the defense to be considered successful, the attack success rate should be brought close to 0 without a substantial loss in model accuracy. We present the hyper-parameters of the targeted embedding attack in Table~\ref{tab:attack_parameters_pruning}.

The pruning defense is not successful on the (DenseNet, CIFAR-10) run, as well as the (VGG, GTSRB) run. Figure \ref{fig:prune_gtsrb_vgg_naive} shows how the model accuracy on the (VGG, GTSRB) setting. Pruning on these settings leads to a fall in model accuracy without a corresponding fall in attack success rate in the earlier stages of pruning, and the model accuracy drops to below 40\% before the backdoor behavior is removed. We thus exclude these settings from our evaluation for feature pruning.

Figure \ref{fig:prune_gtsrb_densenet_naive} shows the model accuracy and attack success rate as pruning is performed for the (DenseNet, GTSRB) setting. The attack is successful in removing the backdoor behavior with a minimal loss in model accuracy. The backdoor behavior is completely removed in exchange for about an 8\% reduction in model accuracy, thus minimal fine-tuning is required to restore the model accuracy. However, the pruning defense is unable to remove the backdoor from the models with targeted adversarial embedding. Figure \ref{fig:prune_gtsrb_densenet_aware} shows the effect of pruning on our defense-aware model. The pruning algorithm selects clean neurons to prune over the backdoor neurons, thus we see the model accuracy fall dramatically when pruning is performed on the network. Furthermore, as the clean neurons are removed, the relative magnitude of the backdoor signals increases, and we see the attack success rate increasing.

\subsection{Results of Adversarial Embedding}
\begin{figure}[t]
	\includegraphics[width=\linewidth]{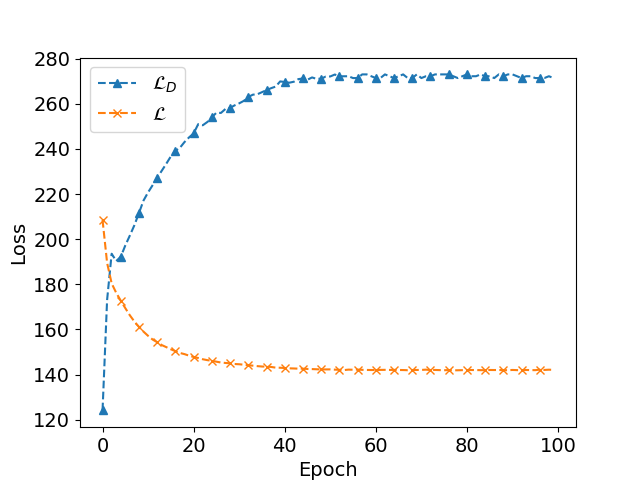}
	\caption{The discriminator loss $\mathcal{L}_{D}$ (Equation \ref{eqn:discriminator_loss}) and the loss $\mathcal{L}$ of the poisoned model (Equation \ref{eqn:adversarial_loss}) versus the epoch when adversarial training is performed. The losses converge to a constant value as the epochs increases.}
	\label{fig:adversarial_losses}
\end{figure}

We then perform our adversarial embedding attack on the baseline model. The parameters of our attack are recorded in Table \ref{tab:attack_parameters}. 30 epochs is sufficient for the attack to mitigate the dataset filtering defenses based on activation clustering and spectral signatures, but 1000 epochs was needed to evade the pruning defense. We use a fully connected discriminator with a leaky ReLU activation and negative slope of 0.2, as well as batch norm in each of the layers. The architecture of the discriminator is described in Table \ref{tab:discriminator_architecture}. Noise sampled from a Gaussian distribution $\mathcal{N}(0, \sigma^2)$ is added to the discriminator inputs. $\sigma$ begins at 0.1, and is decreased by a factor of 10 each epoch. We use a weight decay of 0.9. Figure \ref{fig:adversarial_losses} shows how the losses of the discriminator and the poisoned model converge to their final values as the embedding attack runs.

\begin{figure}[t]
	\centering
	\includegraphics[width=\linewidth]{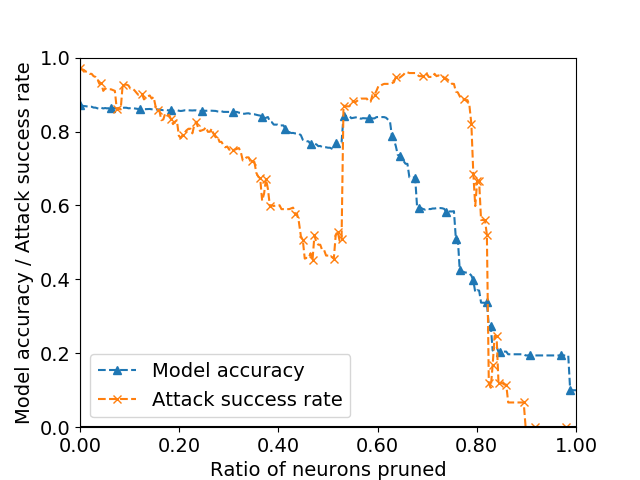}
	\caption{Attack success rate and model accuracy versus the ratio of neurons pruned when pruning \protect{\cite{wang2019neural}} is performed on the model with adversarial embedding (VGG, CIFAR-10). Pruning completely removes the backdoor behavior at the expense of bringing the model accuracy down to 20\%.}
	\label{fig:prune_vgg_hide_cluster}
\end{figure}

Figure \ref{fig:prune_vgg_hide_cluster} shows effect of pruning on the model accuracy and the attack success rate of the VGG model adversarially trained on the CIFAR-10 dataset. While the model is noticeably less robust to the pruning defense than the one trained with targeted adversarial embedding, the model still preserves a significant percentage of the attack success rate as the model accuracy falls. The defense is no longer able to completely remove the backdoor behavior from the model without requiring significant retraining to restore the original model accuracy.

\begin{figure}[t]
	\begin{subfigure}{\linewidth}
		\centering
		\includegraphics[width=\linewidth]{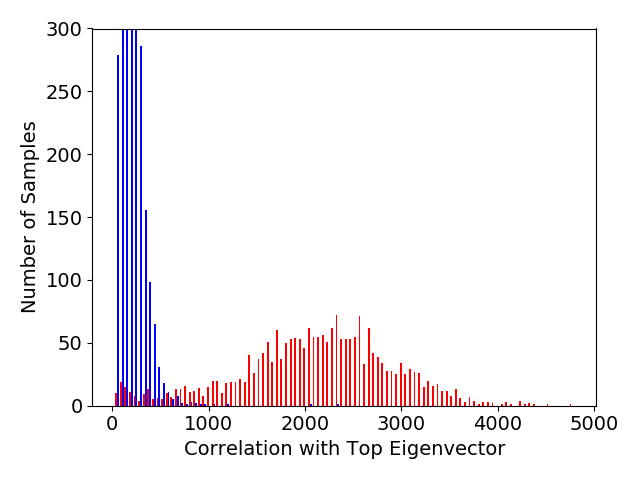}
		\subcaption{Baseline model. The poisoned representations have a significantly higher correlation than the clean representations, and thus can be filtered by removing samples with the top correlations.}
		\label{fig:svd_naive}
	\end{subfigure}
	
	\begin{subfigure}{\linewidth}
		\centering
		\includegraphics[width=\linewidth]{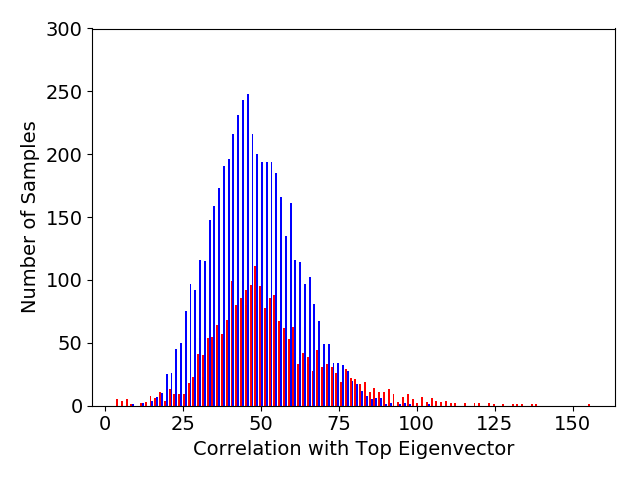}
		\subcaption{Model with adversarial embedding. The poisoned and clean representations have similar distributions of correlations, and thus filtering the samples with top correlations removes an equal portion of clean and poisoned samples.}
		\label{fig:svd_aware}
	\end{subfigure}
	\caption{Correlation of latent representations of all inputs in the training dataset with the top eigenvector for the data filtering defense based on spectral signatures \protect{\cite{tran2018spectral}}. The representations of poisoned inputs are in red.}
\end{figure}

We next evaluate the dataset filtering defense based on spectral signatures \cite{tran2018spectral}. Table \ref{tab:results_robust_statistics} shows the number of poisoned training samples present in the training set before and after the defense is carried out. We use an $\epsilon$ equal to the percentage of poisoned samples in the backdoor target label, so a total of $(1.5\cdot \epsilon)\%$ of samples with the given label are removed. The defense works well on the baseline model, in most cases removing all but less than 2\% of poisoned samples within the training dataset. However, our adversarial embedding of the backdoor successfully mitigates the detection algorithm, leaving more than half of the poisoned samples in the training dataset. We retrain models on the filtered training datasets, and record the attack success rate on the new models in Table \ref{tab:results_robust_statistics}. The defense is successful on the baseline models, and the attack success rate on the retrained model is low (0\% to 1.5\%), but the retrained models from the model with adversarial embedding retains the backdoor behavior, and we see an attack success rate of above 90\%.

To understand the attack, we analyze the correlations of the latent representations with the top eigenvector produced from singular value decomposition. This is the defender's mechanism to remove poisoned samples from the dataset. Figure~\ref{fig:svd_naive} shows the correlations for the baseline model with the (VGG, CIFAR-10) setting. The distribution of correlations of the representations for the poisoned inputs (in red) are noticeably different from that of the clean inputs (in blue), with the representations of the poisoned inputs having much higher correlations. A large fraction of poisoned inputs can thus be removed by removing the inputs whose latent representations have the highest correlations.  However, the {\bf latent representations yield a similar distribution of correlations for the model with adversarial embedding}, as shown in Figure \ref{fig:svd_aware}. The poisoned inputs in the dataset no longer exhibit a larger correlation, and removing inputs with the highest correlations is not sufficient to remove the majority of the poisoned inputs from the training dataset, as poisoned and clean samples are both removed at the same rate.

\begin{figure}[t]
	\begin{subfigure}{\linewidth}
		\includegraphics[width=\linewidth]{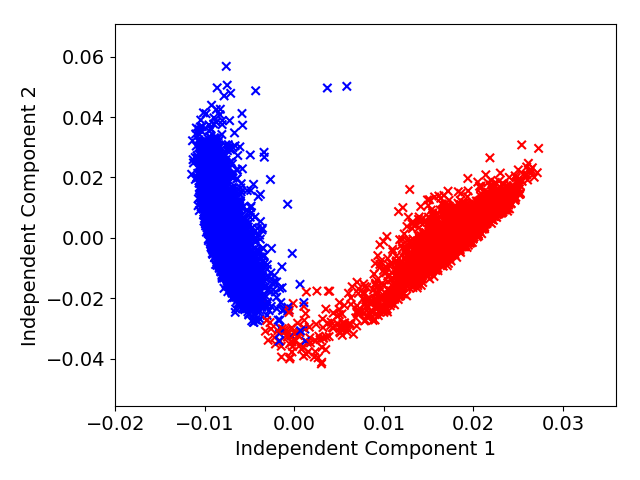}
		\subcaption{Baseline model. The representations of poisoned inputs (in red) form a distinct separate cluster from the clean inputs, and thus are easily separated by k-means clustering.}
		\label{fig:cluster_naive}
	\end{subfigure}
	
	\begin{subfigure}{\linewidth}
		\includegraphics[width=\linewidth]{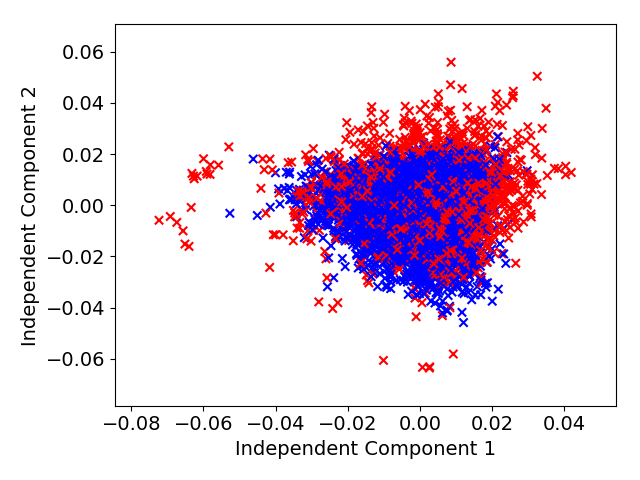}
		\subcaption{Model with adversarial embedding. The representations of poisoned inputs (in red) have a similar distribution as those of clean inputs, thus both clusters formed by k-means clustering contain a significant number of poisoned samples.}
		\label{fig:cluster_aware}
	\end{subfigure}
	\caption{Latent representations of all inputs in the training dataset projected onto their top two independent components for the data filtering defense using activation clustering~\protect{\cite{chen2018detecting}}. The poisoned inputs are depicted in red.}
\end{figure}

We now evaluate the dataset filtering defense based on activation clustering, proposed by Chen et al. \cite{chen2018detecting}. The defense consists of two main steps -- clustering, and performing additional detection steps on the resultant clusters. We notice that the detection methods assume a clean split between the poisoned samples and the clean samples. For example, the exclusionary reclassification step assumes that the newly trained model does not contain the backdoor behavior, and classifies the poisoned samples as their true class. Thus, we compute the quality of the clustering step using the adjusted Rand index to evaluate the quality of the clustering. The perfect separation of poisoned and clean samples will give a score of 1, whereas random separation of the samples will give a score of 0. We also retrain the model on the filtered dataset assuming the best resultant cluster is selected for removal, i.e. the resultant cluster with the highest fraction of poisoned samples is removed. In the clustering step of the defense, we perform independent component analysis to reduce the latent representations to 10 independent components, before running k-means clustering with $k=2$.

\begin{table*}[h]
	\centering
	\begin{tabular}{llcccccc}
		\toprule
		Model & Dataset & $\epsilon$ & 
		\# Pois & 
		\vtop{\hbox{\strut \# Pois Left}\hbox{\strut (Baseline)}} & 
		\vtop{\hbox{\strut \# Pois Left}\hbox{\strut (Adversarial Embedding)}} & 
		\vtop{\hbox{\strut Attack Success}\hbox{\strut (Baseline)}} & 
		\vtop{\hbox{\strut Attack Success}\hbox{\strut (Adversarial Embedding)}} \\
		\midrule
		DenseNet & C10 & 31 & 2241 & 37 (1.7\%) & 1406 (63.0\%) & 1.1\% & 97.3\%\\
		DenseNet & GTSRB & 46 & 1844 & 27 (1.5\%) & 1263 (68.5\%) & 0.0\% & 89.9\%\\
		VGG & C10 & 31 & 2241 & 54 (2.4\%) & 1050 (46.9\%) & 1.5\% & 97.3\%\\
		VGG & GTSRB & 46 & 1844 & 11 (0.6\%) & 1004 (54.4\%) & 0.3\% & 90.3\%\\
		\bottomrule
	\end{tabular}
	\caption{Number of poisoned samples left in the dataset, as well as the attack success rate on the retrained models for the dataset filtering defense based on spectral signatures \protect{\cite{tran2018spectral}}. The defense removes most poisoned samples from the baseline model, but fails to do so for the model with adversarial embedding.}
	\label{tab:results_robust_statistics}
\end{table*}

\begin{table*}[h]
	\centering
	\begin{tabular}{llccccc}
		\toprule
		Model & Dataset & \vtop{\hbox{\strut Rand Index}\hbox{\strut (Baseline)}} & \vtop{\hbox{\strut Rand Index}\hbox{\strut (Adversarial Embedding)}} & 
		\vtop{\hbox{\strut Attack Success}\hbox{\strut (Baseline)}} & 
		\vtop{\hbox{\strut Attack Success}\hbox{\strut (Adversarial Embedding)}} \\
		\midrule
		DenseNet & CIFAR-10 & 0.979 & 0.182 & 1.7\% & 76.4\% \\
		DenseNet & GTSRB & 0.997 & 0.271 & 0.0\% & 91.4\% \\
		VGG & CIFAR-10 & 0.998 & $6.31 \times 10^{-4}$ & 1.9\% & 96.2\% \\
		VGG & GTSRB & 0.997 & 0.642 & 1.1\% & 74.3\% \\
		\bottomrule
	\end{tabular}
	\caption{Adjusted Rand indexes of the clusters formed from k-means clustering with $k=2$ in the defense based on activation clustering \protect{\cite{chen2018detecting}}, as well as the attack success rate on the models retrained on the filtered dataset. A Rand index close to 1 represents a good separation of clean and poisoned inputs in the clustering step, while a Rand index close to 0 represents poisoned inputs are present in both sets. Adversarial embedding is able to significantly reduce the quality of clustering, and thus the retrained model still contains the backdoor behavior.} 
	\label{tab:results_clustering}
\end{table*}

Table~\ref{tab:results_clustering} shows the adjusted Rand scores of the clusters formed, as well as the attack success rate on the retrained model for both the baseline, and the model with adversarial embedding. The defense works well on the baseline model, attaining almost perfect clustering of the clean and poisoned samples (adjusted Rand index $>$ 0.95), and the attack success rate on the retrained models is very low (0\% to 1.9\%). However, our attack successfully bypasses this defense technique, yielding a lower resultant adjusted Rand index after the clustering step. This signifies that a large number of poisoned samples is present in both resultant clusters. Thus, the model retrained on the resultant filtered dataset still contains the backdoor behavior, and exhibits a high attack success rate of above 75\%.

To understand why our attack evades the defense, we analyze the latent representations of the VGG model trained on the CIFAR-10 dataset. Figure \ref{fig:cluster_naive} shows the latent representations of all inputs by the baseline model, projected onto the top two independent components. The poisoned inputs (in red) form a noticeably disjoint cluster from the clean inputs (in blue). Thus, k-means clustering can separate the poisoned inputs from the clean inputs well, and most of the poisoned samples can be removed from the training dataset by excluding the right cluster. Figure \ref{fig:cluster_aware} shows the latent representations for the model with adversarial embedding, similarly projected onto the top two independent components. The {\bf adversarial training causes the poisoned and clean latent representations to overlap significantly due to a convergence in their distributions}, so k-means clustering is unable to separate the poison inputs from the clean inputs well. Thus, removing either of the resultant k-means clusters still leaves a significant number of poisoned sample in the training dataset, and the retrained model will still contain the backdoor.

\section{Related Work}

Backdoors in deep learning networks is a topic of growing interest. Many studies explore backdoor injection through data poisoning, where the attacker injects maliciously crafted input and label samples into the training dataset \cite{gu2017badnets, chen2017targeted, shafahi2018poison}. These poisoned inputs are typically images with the backdoor trigger superimposed, and the target label altered. It has been shown that a small number of poisoned data points (50 samples) is needed to introduce a backdoor with a high attack success rate (above 90\%) \cite{chen2017targeted}. It has been shown that the introduced backdoors persist after the model is repurposed \cite{gu2017badnets}. Thus, malicious behavior injected in models up the supply chain can propagate to downstream models, even if the retraining is done with a clean dataset. Many approaches to data poisoning lead to a visible trigger superimposed on the poisoned images, which makes malicious training data apparent to human eyes, but it has been shown to be possible to generate poison data that looks clean with unaltered labels \cite{shafahi2018poison}.

Backdoor attacks that do not rely on the attacker having access to the training data have also been devised. In the federated learning setting, the attacker has the ability to broadcast weight updates to the other parties. This ability can be exploited to broadcast weight updates that introduce backdoor behavior to the models that receive the update \cite{bagdasaryan2018backdoor}. Federated learning settings with secure aggregation are especially susceptible to this attack as the individual weight updates cannot be inspected. Furthermore, the algorithm that generates the weight updates can also take into account the anomaly detection technique, in order to bypass the defense when secure aggregation is not used. Various studies have also worked on making distributed learning settings that converge to a useful model despite the presence of Byzantine workers \cite{xie2018generalized,yin2018byzantine,blanchard2017machine}. The guarantees that these robust algorithms provide, however, are shown to be insufficient \cite{mhamdi2018hidden,baruch2019little}.

In response to the backdoor attacks that have been devised, there have been several papers that aim to remove backdoor behavior, besides the ones mentioned in this paper. Liu et al., 2018 \cite{liu2018fine} take a similar approach of pruning neurons based on the latent representations of a known clean set of inputs, operating under the assumption that the backdoor neurons are dormant for clean inputs. Since it is possible to train a model to be robust to this pruning, Liu et al. recommend a combination of pruning and fine-tuning of the model to remove the backdoor behavior. Some backdoor mitigation techniques work on the input space of the model instead of the latent space, relying on either training a model to identify anomalous inputs, or to remove anomalous features in inputs before feeding to the model \cite{liu2017neural}.

Besides backdoor attacks, there have been many studies on adversarial machine learning where the discontinuous input-output mappings of models are exploited to generate adversarial examples~\cite{szegedy2013intriguing,goodfellow2014explaining}. It has also been shown to be possible to generate adversarial images by perturbing the color space of the image, thus preserving the smoothness of the image, in order to evade detection methods that rely on the abrupt pixel changes found in many adversarial image detection methods \cite{hosseini2018semantic}.

\section{Conclusions}

We have designed a novel backdoor embedding attack that successfully bypasses several prior backdoor detection algorithms.  While backdoor detection using the learned latent representations greatly reduces the dimensionality and thus complexity of the defense techniques, we have shown that a sophisticated attacker is easily able to hide the signals of the backdoor images in the latent representation, rendering the defense algorithms ineffective. 
	
\section*{Acknowledgments}

This work is supported by the NUS Early Career Research Award (NUS ECRA) by the Office of the Deputy President, Research \& Technology (ODPRT), grant number NUS\_ECRA\_FY19\_P16.

\end{document}